\documentclass[]{article} 

\usepackage{amssymb}
\usepackage{amsmath}
\usepackage{graphicx}
\usepackage{subfigure}
\usepackage{makeidx}
\usepackage{multicol}
\usepackage[export]{adjustbox}
\usepackage[justification=centering]{caption}
\usepackage{chngcntr}
\counterwithout{figure}{section}
\usepackage{float}
\usepackage{threeparttable}
\usepackage{xcolor} 

\usepackage[utf8]{inputenc}

\usepackage{bm}
\usepackage{booktabs}
\usepackage{multirow}
\usepackage{caption}
\usepackage{tabularx}

\usepackage{xcolor}
\def\RED#1{\textcolor{black}{#1}}

\def\B#1{#1}
\def\G#1{#1}

\usepackage{fancyhdr}
\let\origtitle\title 
\renewcommand{\title}[1]{\lfoot{\textit{#1}}\origtitle{\textbf{#1}}}
\renewcommand{\sectionmark}[1]{\markboth {}{}}

\date{}

\title{Multimodal Foundation Model for Cross-Modal Retrieval and Activity
Recognition Tasks}

\begin{document}
\maketitle
\thispagestyle{fancy}
\centering

\author{Koki Matsuishi
\footnote{koki.matsuishi587@mail.kyutech.jp}, 
 Kosuke Ukita
\footnote{ukita.kosuke299@mail.kyutech.jp}
 Tsuyoshi Okita \footnote{tsuyoshi.okita@gmail.com}}\\
\thanks{$^1$$^2$$^3$Kyushu Institute of Technology}

\abstract{
In recent years, the widespread adoption of wearable devices has highlighted the growing importance of behavior analysis using IMU. While applications span diverse fields such as healthcare and robotics, recent studies have increasingly focused on multimodal analysis, in addition to unimodal analysis. 
Several studies have proposed multimodal foundation models that incorporate first-person video and text data; however, these models still fall short in providing a detailed analysis of full-body human activity.
To address this limitation, we propose Activity Understanding and Representations Alignment - Multimodal Foundation Model (AURA-MFM), a foundational model integrating four modalities: third-person video, motion capture, IMU, and text. 
By incorporating third-person video and motion capture data, the model enables a detailed and multidimensional understanding of human activity, which first-person perspectives alone fail to capture.
Additionally, a Transformer-based IMU encoder is employed to enhance the model's overall performance. 
Experimental evaluations on retrieval and activity recognition tasks demonstrate that our model surpasses existing methods.
Notably, in the zero-shot classification for action recognition, our method achieved significantly higher performance, with an F1-score of 0.6226 and an accuracy of 0.7320, whereas the existing method recorded an F1-score of 0.0747 and an accuracy of 0.1961.
}

\begin{figure}[h]
    \centering
    \includegraphics[scale=0.35]{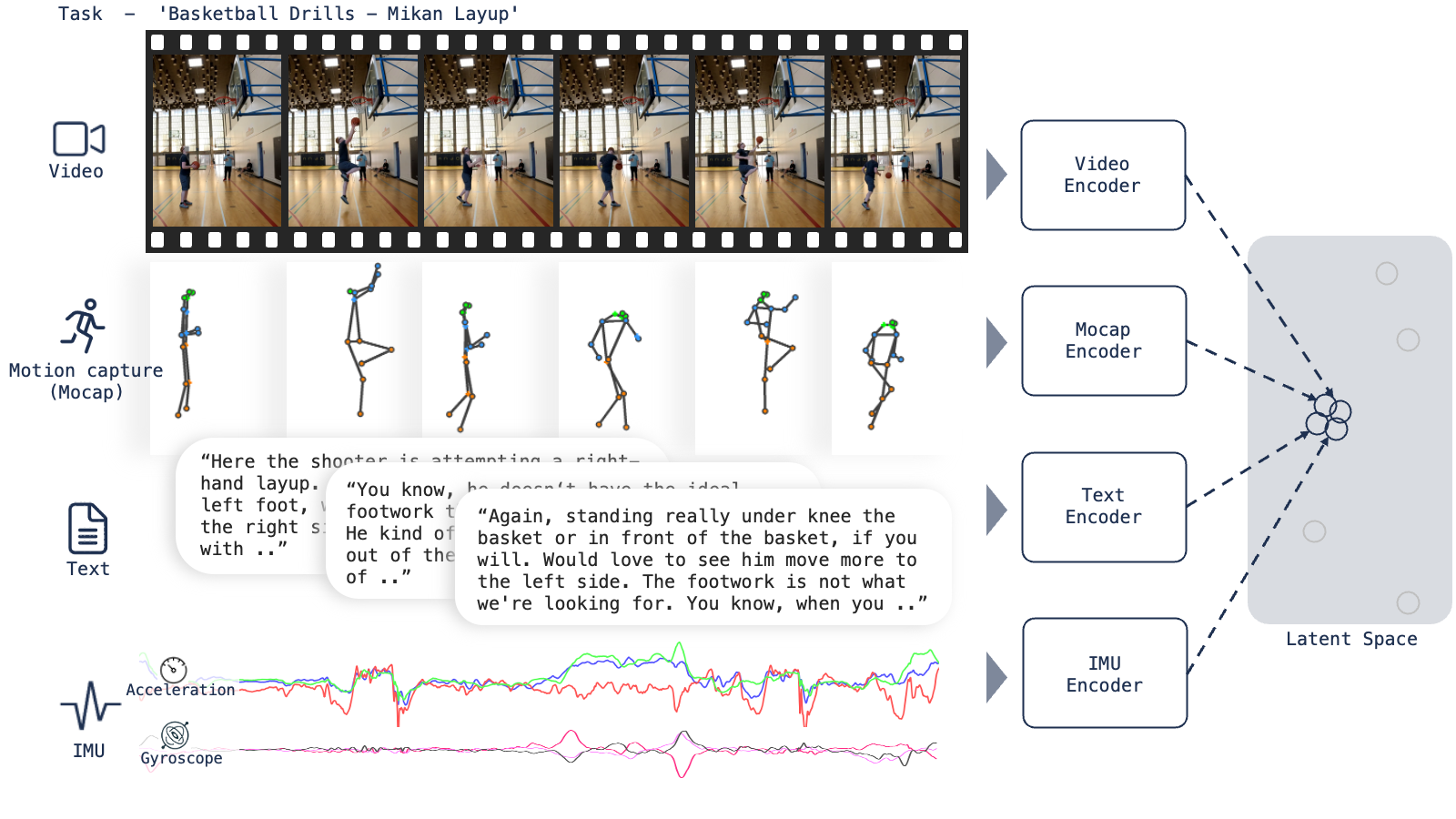}
    \caption{\textbf{AURA-MFM}: A multimodal foundational model that integrates four modalities—text, motion capture, third-person perspective videos, and IMU (accelerometer and gyroscope). Each modality is processed by a dedicated encoder, and their synchronized embeddings are mapped into a shared latent space. By minimizing the distance between embeddings from different modalities, AURA-MFM aims to enhance the understanding of human activities.}
    \label{fig:overview}
\end{figure}
\section{Introduction}
In recent years, the widespread adoption of smart glasses and wearable devices has increased interest in human activity analysis using built-in sensors such as accelerometers and gyroscopes (IMUs). 
This technology has demonstrated significant potential across various domains, including healthcare, sports, rehabilitation, and robotics. 
While prior research has primarily focused on either IMU data or video in isolation, recent studies increasingly integrate multiple modalities to develop more robust and versatile multimodal models.
Within this field, novel approaches leveraging IMU data have been gaining attention, with several noteworthy studies reported (\cite{bhalla2021imu2doppler}, \cite{kwon2020imutube}, \cite{deldari2022cocoa}).

In multimodal foundation models, IMU data must be aligned with the representational space of other modalities such as video and text. One approach to achieving this is to encode IMU data into a representational form using an RNN (Recurrent Neural Network)-based IMU encoder, as demonstrated in IMU2CLIP \cite{moon2023imu2clip}. 
However, RNNs are known to struggle with vanishing gradient issues and inefficiencies, especially for processing extended time-series data or complex motion information.
The adoption of this approach in IMU2CLIP likely stemmed from the relatively low emphasis placed on IMU data, serving as a temporary solution rather than an optimal choice. This limitation is particularly evident in cross-modal contrastive learning, where the expressive power of RNN-based models has been criticized.

An alternative and well-established approach is to adopt a Transformer\cite{vaswani2017attention}-based architecture \cite{vaswani2017attention}. By leveraging the self-attention mechanism, Transformers can efficiently capture long-range dependencies in time-series data and integrate critical information across the entire sequence. This capability makes them well-suited for learning refined representations of IMU data. Moreover, Transformers are particularly effective in forming a consistent representational space across multiple modalities, as they can capture not only local dependencies but also global correlations. This characteristic makes them a promising choice for cross-modal contrastive learning.
\RED{Conventional multimodal foundation models face certain limitations in handling ubiquitous tasks such as human activity recognition. For example, most methods that utilize video as a modality—such as IMU2CLIP—primarily rely on first-person perspective videos. However, first-person videos are limited to the viewpoint of the camera wearer, making it difficult to capture full-body movements and interactions with the surrounding environment in detail. To achieve high performance across a wide range of downstream ubiquitous tasks, it is essential to choose modalities that can provide sufficient information for detailed analysis of human motion. This important consideration has not been adequately addressed in existing multimodal foundation models for ubiquitous tasks.
In response to these challenges, we propose \textbf{AURA-MFM} (\textbf{A}ctivity \textbf{U}nderstanding and \textbf{R}epresentations \textbf{A}lignment - \textbf{M}ultimodal \textbf{F}oundation \textbf{M}odel), a novel multimodal foundation model. As shown in Figure 1, AURA-MFM integrates four distinct modalities: IMU data, third-person perspective videos, text, and motion capture data. Within a contrastive learning framework, the model aims to enhance understanding and representation of human activity.} 

Each of these modalities captures complementary aspects of human activity, enabling a more comprehensive representation of movements and actions. By leveraging dedicated encoders for each modality, AURA-MFM processes and aligns multimodal information in a shared latent space, facilitating a deeper and more structured understanding of human behavior across different sensory inputs.
\RED{The IMU encoder adopts a Transformer-based architecture, enabling more refined representation learning compared to conventional RNN-based encoders. Moreover, by introducing third-person perspective videos and motion capture as new modalities, AURA-MFM allows for a more detailed analysis of full-body human movements. To the best of our knowledge, this is the first study to introduce motion capture and third-person view videos for detailed activity understanding in multimodal foundation models.} The significance of this study lies in its ability to model human activities and movements more precisely and comprehensively by combining multiple data modalities.

The contributions of this paper are as follows:
\begin{itemize}
\item \RED{We propose a new multimodal foundation model (AURA-MFM) that integrates four modalities: third-person perspective videos, text, motion capture, and IMU data.  Third-person videos and motion capture, which were lacking in previous models, are introduced to enable more detailed understanding of human activity.}
\item \RED{We propose a Transformer-based IMU encoder, which learns more efficient latent representations compared to a RNN-based IMU encoder.}
\item \RED{Through experiments, we demonstrate that AURA-MFM outperforms the previous model in human activity recognition tasks.}
\end{itemize}

The remainder of this paper is organized as follows. Section \ref{section:relatedwork} provides an overview of related work and fundamental background concepts relevant to this study. Additionally, key theoretical foundations that support the proposed framework are outlined. In Section \ref{section:methods}, the AURA-MFM multimodal foundational model is introduced. Section \ref{section:experiments} presents the experimental evaluation, including details on datasets, preprocessing techniques, evaluation metrics, and results. Finally, Section \ref{section:conclusion} concludes the paper by summarizing the key findings and contributions.

\section{Related Work} \label{section:relatedwork}
\subsection{Self-supervised Learning}
Self-supervised learning is a technique for learning feature representations from large amounts of unlabeled data. It has recently gained significant attention in deep learning. Self-supervised learning generates pseudo-labels from the structure of the data itself and uses them as supervisory signals for training. This approach enables the acquisition of powerful feature representations without the need for large-scale labeled datasets, unlike traditional supervised learning. A pre-trained model built through self-supervised learning can be applied to downstream tasks such as classification using a small amount of labeled data.

Representative methods include various pretext tasks designed to facilitate self-supervised learning. For instance, the Jigsaw Puzzle Solving method \cite{10.1007/978-3-319-46466-4_5} shuffles image patches and trains a model to recover the correct arrangement, thereby capturing spatial relationships. Similarly, the Rotation Prediction method \cite{conf/iclr/GidarisSK18} rotates images by fixed angles and trains a model to predict the rotation angle, learning spatial and semantic representations.Another notable approach is BERT \cite{devlin-etal-2019-bert}, which is based on autoregressive prediction. BERT utilizes a Masked Language Model (MLM) to predict masked words, thereby capturing context-aware representations. Self-supervised learning has been widely applied in various fields, including natural language processing and computer vision. It is particularly effective in problems where large-scale data annotation is challenging.\\
\textbf{Contrastive Learning} \\
Contrastive Learning is a self-supervised learning approach that aims to learn similarities between data instances. It works by defining "positive pairs" as different augmented versions of the same data and "negative pairs" as unrelated data. The model is trained to pull positive pairs closer in the feature space while pushing negative pairs apart.
Prominent methods include SimCLR \cite{10.5555/3524938.3525087} and MoCo \cite{He_2020_CVPR}. SimCLR applies different augmentations to image data and learns representations by treating them as belonging to the same instance. MoCo, on the other hand, employs a dynamic queue to maintain a large number of negative examples, enabling more stable contrastive learning even with a small batch size. Contrastive learning has gained significant attention in unsupervised learning frameworks, particularly in image recognition and natural language processing. By leveraging the structure of data, it enables the acquisition of high-quality feature representations.\\
\subsection{Foundation Models}
A Foundation Model \cite{bommasani2022opportunitiesrisksfoundationmodels} is a general-purpose machine learning model pre-trained on a massive dataset and adaptable to a wide range of downstream tasks. These models are utilized across diverse domains, including natural language processing (NLP) and computer vision.

In the field of NLP, Transformer-based models such as BERT\cite{devlin-etal-2019-bert} and GPT\cite{NEURIPS2020_1457c0d6} are widely recognized. Through pre-training, these models acquire the ability to understand the meaning of sentences and generate text, and are then adapted to various tasks, including translation and question answering, via fine-tuning. In the field of computer vision, the Vision Transformer\cite{dosovitskiy2021an} has gained attention as an alternative approach to CNNs. By processing images in patches, it has achieved high-accuracy image classification.\\
\textbf{CLIP} \\
In recent years, multimodal foundation models that handle data integrally across modalities like images and language have been gaining attention. A representative example of this is CLIP\cite{radford2021learning}.
CLIP is a multimodal foundation model proposed by OpenAI. CLIP is trained using a large collection of image-text pairs, allowing it to capture the relationships between text and images with high accuracy.
During training, CLIP adopts a contrastive learning approach where an image and its corresponding caption form a "positive pair," while combinations with other captions are treated as "negative pairs." This enables CLIP to perform zero-shot learning, where it can retrieve images based on textual descriptions—even for images it has never seen during training. For example, when given the text "a dog image," CLIP can correctly identify relevant images even if they were not part of the training dataset.

CLIP has numerous applications, including zero-shot classification, image retrieval, and text-to-image generation. Compared to traditional supervised learning models, it offers greater flexibility and robustness, making it particularly useful in scenarios where labeled data is scarce. As a breakthrough in integrating vision and language understanding, CLIP represents a significant advancement.\\
\textbf{Multimodal Foundation Models for Sensor Data}\\
The increasing availability of sensor data has spurred interest in multimodal approaches that leverage various sensor modalities, including IMU (Inertial Measurement Unit) data. Several recent studies have explored the integration of IMU data with other modalities using foundation models. IMU2CLIP \cite{moon2023imu2clip} is a notable example that connects IMU data to CLIP, establishing a link between wearable sensor data and natural language, as well as first-person perspective videos.  IMU2CLIP demonstrated the potential of mapping IMU data to textual and video representations.  As a downstream application, IMU2CLIP proposed using IMU data to retrieve corresponding media, such as egocentric videos or text narrations, and showed performance improvements in activity recognition tasks. However, IMU2CLIP utilizes an RNN-based IMU encoder, and the effectiveness of Transformer-based encoders for IMU data within this framework remains unexplored. Furthermore, its current scope does not extend to integrating third-person perspective videos or motion capture data.

IMU2Doppler \cite{bhalla2021imu2doppler} learns to map human activity representations from IMU data to millimeter-wave radar sensor representations. They improved human activity recognition performance using only 15 seconds of labeled Doppler radar data. \cite{tong2021zero} proposed a method to directly transfer video representations to IMU representations. By extracting action representations from a pre-trained video model and learning a projection from the IMU representation space to the video representation space, they enhanced recognition performance for activities that IMU data alone could not recognize. COCOA \cite{deldari2022cocoa} introduced a novel objective function to compute cross-modal correlations while minimizing similarities between irrelevant instances. This self-supervised model learned high-quality representations from multi-sensor data, demonstrating superior classification performance compared to baselines. ImageBind \cite{girdhar2023imagebind} introduced six modalities: images, text, audio, depth, thermal, and IMU data. Instead of providing paired data for all modalities, it trained encoders only between images and other modalities, demonstrating that the image modality can serve as a foundational representation connecting all other modalities.

\subsection{Ego-Exo4D} 

\B{The experiments in IMU2CLIP used the Ego4D dataset which is one motivation in our study. Thus, this fact lead to the several challenges that we invoke in this paper.}
Specifically, the use of first-person perspective videos limits the viewpoint to the camera wearer, making it difficult to comprehensively capture full-body movements and interactions with the surrounding environment. Additionally, due to the use of datasets with limited viewpoints, the integration with IMU data tends to restrict the scope of motion analysis.

In contrast, this study focuses on third-person perspective videos and motion capture data, aiming for a more detailed analysis of full-body human movements that first-person perspectives alone cannot fully capture. To achieve this, we adopt the Ego-Exo4D dataset \cite{Grauman_2024_CVPR}, which integrates both first-person (ego) and third-person (exo) perspectives, providing a new foundation for a multi-view understanding of human activity. The Ego-Exo4D dataset includes motion capture data that captures full-body movements, complementing the perspectives and features that IMU2CLIP and the conventional Ego4D dataset \cite{Ego4D2022CVPR} could not provide.

\B{Before the era of Ego4D and Ego-Exo4D datasets, there have been various kinds of multimodal dataset as well. Among them, one dataset was Berkeley MHAD dataset \cite{BerkeleyMHAD}. This MHAD dataset has mocap, IMU, and video. This dataset does not have text modalities. Unfortunately, this MHAD dataset cannot be accessed. Similarly, many multimodal dataset developed at the same era with Berkeley MHAD dataset, which was more than 5 years ago which means that the day before CLIP appears, does typically lack the modality of text. CLIP and other recent LLMs and foundation models connects the dataset of images and text. In this reason, among various multimodal dataset, we focus solely on Ego-Exo4D dataset as well as Ego dataset in this paper.}

\begin{figure}[t]
    \centering
    \includegraphics[scale=0.5]{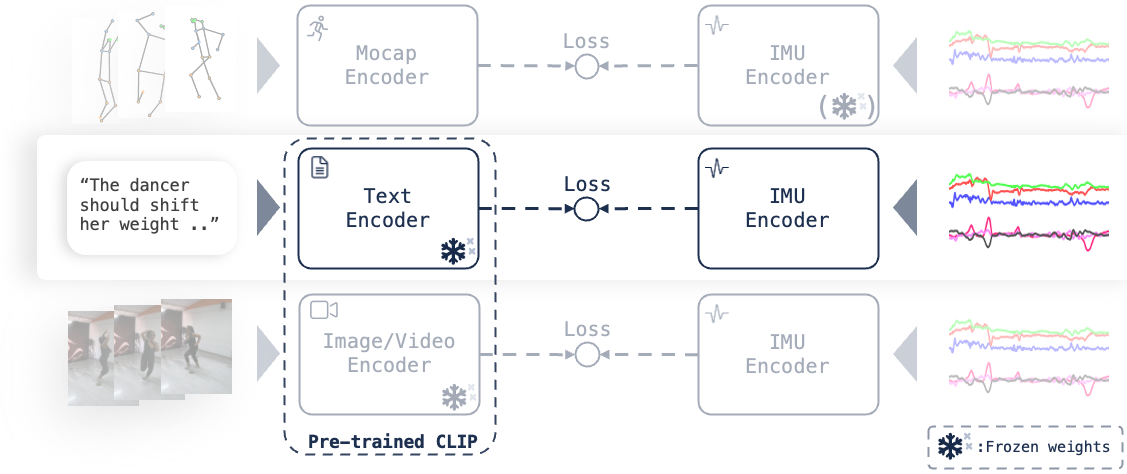}
    \caption{\textbf{Cross-Modal Contrastive Learning}: We perform cross-modal contrastive learning between IMU data and other modalities. The text encoder and video encoder use the publicly available pre-trained CLIP parameters, and their parameters are fixed during training. Although the figure specifically illustrates training between text and IMU, the model is trained using three types of contrastive learning: video and IMU, motion capture and IMU, and text and IMU. This approach enables the construction of a foundation model that integrates multiple modalities.}
    \label{fig:foundationmodel}
\end{figure}

\section{Proposed Method} \label{section:methods}

In Section \ref{section:methods}, we present our proposed approach, AURA-MFM, in detail. This section provides a comprehensive explanation of how our AURA-MFM is constructed and how it integrates different modalities for enhanced human activity understanding.
Section \ref{section:method1} focuses on the core architecture of AURA-MFM, describing the methodology used to create a shared latent space where multiple modalities—third-person perspective videos, motion capture data, IMU signals, and textual descriptions—are aligned. 
Section \ref{section:method2} introduces the newly developed IMU encoder and Mocap encoder, which are designed to enhance the feature extraction process from their respective data streams. 

\subsection{Proposed Multimodal Foundation Model (AURA-MFM)} \label{section:method1}

Training an IMU encoder using contrastive learning with a pre-trained CLIP model \cite{radford2021learning}, as introduced by IMU2CLIP \cite{moon2023imu2clip}, represents a significant advancement in the integration of IMU data into vision-language models. Inspired by this approach, we extend the framework beyond IMU by incorporating third-person perspective videos and motion capture data. This enables the development of AURA-MFM, a new multimodal foundation model that integrates and aligns four modalities: video, text, motion capture, and IMU. By doing so, our model aims to create a unified representation space where embeddings from different sensor data sources can be directly compared and leveraged for downstream tasks requiring a comprehensive understanding of human activity.

As illustrated in Figure \ref{fig:foundationmodel}, our approach is based on cross-modal contrastive learning, where the IMU encoder is trained by aligning its representations with those of other modalities. The encoders for text and video utilize publicly available pre-trained CLIP parameters, which remain fixed during training. This ensures that the latent space retains the rich semantic structures already learned by CLIP while enabling effective integration of additional data.
Figure \ref{fig:foundationmodel} specifically illustrates contrastive training between text and IMU, but our model is designed to support multiple types of contrastive learning: video and IMU, motion capture and IMU, and text and IMU. By leveraging these diverse contrastive pairs, AURA-MFM constructs a shared multimodal representation space where different sensory inputs can be meaningfully aligned. 
This methodology facilitates the construction of a robust multimodal foundation model capable of capturing complex relationships among various human activity data sources.

To incorporate motion capture and IMU modalities into this space, a progressive training strategy is employed. First, we train the IMU encoder using contrastive learning with text or video while keeping the parameters of the text and video encoders fixed. Once the IMU encoder has learned meaningful representations aligned with CLIP, we introduce the motion capture encoder and train it in conjunction with the IMU encoder. This stepwise approach ensures that the representations of motion capture data and IMU signals are smoothly integrated into the CLIP-aligned latent space without disrupting the semantic structures established in earlier stages.
Additionally, experiments are conducted in which the motion capture and IMU encoders are trained independently, without relying on CLIP representations.

\subsection*{Cross-Modal Contrastive Learning}

\B{
For the loss function, we adopt InfoNCE \cite{oord2018representation}, as it was identified as the most effective among InfoNCE, Triplet Loss \cite{schroff2015facenet}, and Mean Squared Error (MSE) in the investigation conducted by IMU2CLIP \cite{moon2023imu2clip}. \\
\textbf{InfoNCE}: A contrastive loss function commonly used in self-supervised learning, maximizes the mutual information between positive pairs while minimizing their similarity with negative samples. It has been widely utilized in representation learning to enhance feature discrimination. \\\textbf{Triplet Loss}: On the other hand, Triplet Loss is a metric learning approach that optimizes the embedding space by reducing the distance between an anchor and a positive sample while ensuring a margin-based separation from a negative sample. Although effective in tasks such as face recognition and similarity learning, it heavily depends on the selection of triplets during training. \\
\textbf{MSE}: Lastly, MSE, a standard loss function for regression tasks, calculates the mean squared difference between predicted and true values, making it suitable for scenarios where precise numerical accuracy is required. While Triplet Loss and MSE offer benefits in their respective domains, InfoNCE has been shown to yield superior results in contrastive learning settings, leading to its adoption in our framework.
}

Following \cite{moon2023imu2clip}, given a batch size $B$ of IMU embeddings ($\bm{I}$) and corresponding embeddings from another modality ($\bm{M}$), where $\bm{I} = (\bm{i_1}, \bm{i_2}, ..., \bm{i_B})$ and $\bm{M} = (\bm{m_1}, \bm{m_2}, ..., \bm{m_B})$, the loss for mapping $\bm{I}$ to $\bm{M}$ is defined as Equation (\ref{loss_im}), while the loss for mapping $\bm{M}$ to $\bm{I}$ is defined as Equation (\ref{loss_mi}). Consequently, the final cross-modal loss between $\bm{I}$ and $\bm{M}$ is given by their average, as shown in Equation (\ref{loss}). Here, $\bm{i_i}$ and $\bm{m_i}$ represent encoded embeddings constrained to the unit hypersphere. Since the embeddings are normalized to unit length, their similarity is computed using the inner product between them.

\begin{equation}
\label{loss_im}
L_{\bm{I}\rightarrow \bm{M}} = -\frac{1}{B} \sum_{i=1}^{B} log \frac{\exp(\langle \bm{i_i},\bm{m_i}\rangle)}{\sum_{k=1}^{B} \exp(\langle \bm{i_i},\bm{m_k}\rangle)}
\end{equation}
\begin{equation}
\label{loss_mi}
L_{\bm{M}\rightarrow \bm{I}} = -\frac{1}{B} \sum_{i=1}^{B} log \frac{\exp(\langle \bm{m_i},\bm{i_i}\rangle)}{\sum_{k=1}^{B} \exp(\langle \bm{m_i},\bm{i_k}\rangle)}
\end{equation}
\begin{equation}
\label{loss}
L_{\bm{I}\leftrightarrow \bm{M}} = \frac{1}{2} (L_{\bm{I}\rightarrow \bm{M}} + L_{\bm{M}\rightarrow \bm{I}})
\end{equation}

\subsection{Transformer-Based Encoder Architecture} \label{section:method2}

\begin{figure}[h]
\centering
\includegraphics[scale=0.7]{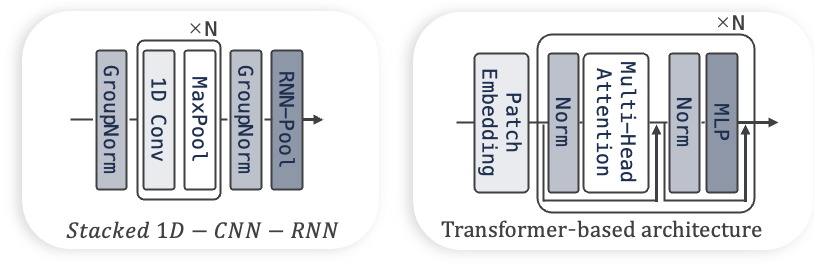}
\caption{Encoder architectures for IMU and Mocap (Left: \textit{Stacked 1D-CNN-RNN} \cite{moon2023imu2clip}, Right: the transformer-based architecture designed for signal processings [anonymous reference])}
\label{fig:architecture}
\end{figure}

\B{Since our approach is based on cross-modal contrastive learning where the IMU encoder is trained by aligning its representations with those of other modalities, the performance of each encoder affects the overall performance.}
IMU2CLIP \cite{moon2023imu2clip} proposed \textit{Stacked 1D-CNN-RNN} as the most effective IMU encoder. In contrast, we adopt the transformer-based architecture designed for signal processing [anonymous reference]. We introduce two modifications to this architecture: adjusting the patch embedding layer size and adding a layer to match the embedding dimensions of CLIP. These modifications enable the use of this architecture as both the IMU encoder and Mocap encoder in our experiments.

Figure \ref{fig:architecture} illustrates the structures of the RNN-based \textit{Stacked 1D-CNN-RNN} (top) and the transformer-based architecture designed for signal processing (bottom). \B{As this is Transformer, we place the self attention mechanism. That is, we place the multi-head attention module before MLP module and after the patch embedding module. The layer normalization module and skip connections are also added as these are invented as the basic architecture of conventional transformer with self-attention mechanism. Note that we deploy the patch embedding part for signal processing [anonymous reference]: this consists of the pretraining objectives of the masked signal model (this term follows from the masked language model in text and the masked image model in image) and the pretext task in signals. In this way, the structure of this Transformer is purely the signal-based T
transformer. This is neither vision transformer nor (language) transformer. 
We developed this transformer for IMU signals. In terms of mocap signals, we adopt the same transformer as well. This is
since we can think that the mocap signal is the multi-dimensional signal of location. This means that the data structure of mocap signals is exactly the same as the IMU signals.}

\B{In terms of other modalities such as image/video and text encoders, we deploy the same encoders as IMU2CLIP since these encoders are already trained on the large-scale data.}

\section{Experiments} \label{section:experiments}
In Section \ref{section:experiments}, we provide a detailed description of the preprocessing steps and experimental results. First, Section \ref{section:datasets} introduces the dataset used in our experiments, explaining its composition and characteristics. Additionally, we describe the preprocessing methods applied to ensure consistency across different modalities, including data normalization, synchronization techniques, and feature extraction procedures.
Next, Section \ref{section:tasks} elaborates on the evaluation methodology, focusing on two key tasks: retrieval and human activity recognition. 
Finally, Section \ref{section:result} presents the experimental results through a combination of figures and tables, illustrating the model’s performance across different tasks. In particular, we analyze the retrieval task on unseen prompts, highlighting AURA-MFM’s ability to generalize beyond the training data. This analysis provides empirical evidence of the model’s robustness and effectiveness in integrating multimodal information.

\subsection{Dataset} \label{section:datasets}

\RED{In this experiment, we primarily use the Ego-Exo4D dataset \cite{Grauman_2024_CVPR}. To evaluate the generalization performance of AURA-MFM, we additionally use the PAMAP2 dataset \cite{6246152} for a downstream task—human activity recognition.} Below, we provide a brief explanation of each modality used in the experiment along with its preprocessing method. 
\B{Table} \ref{dataset} shows the number of dataset samples after preprocessing with the sliding window approach.

\B{In the first line of Table \ref{dataset}, we extract the parallel data between IMU and Mocap with the synchronization of sliding window of 5 seconds. Since the original Ego-Exo4D dataset did not provide such pairs of 5 seconds, these pairs were built by us. Then, we obtained the training data of 2045 pairs, validation data of 108 pairs, and test data of 240 pairs. Similarly, we extract the parallel data between IMU and Video, between IMU and AAD,
between IMU and Expert Commentary, and between IMU and 8 classes. As is explained in the Ego-Exo4D dataset subsection, this data has two kinds of text data. The one is AAD and the other is expert commentary. Both are the detailed text description of the scene or video.}

\begin{table}[h]
    \centering
    \caption{\RED{The number of data after preprocessing the dataset}}
    \label{dataset}

    \begin{tabular}{ccccc} \hline\hline
        Dataset & Data Modarity & Train & Validation & Test \\ \hline
        \multirow{5}{*}{Ego-Exo4D} & IMU $\leftrightarrow$ Mocap (5s) & 2045 & 108 & 240 \\
        &IMU $\leftrightarrow$ Video (5s) & 121k & 16k & 1k \\
        &IMU $\leftrightarrow$ AAD (5s) & 197k & 25k & 1k \\
        &IMU $\leftrightarrow$ Expert Commentary (5s) & 17k & 1k & 1k \\
        &IMU $\rightarrow$ 8 classes (5s) & 3385 & 377 & 418 \\ \hline
        PAMAP2 &IMU $\rightarrow$ 18 classes (5s) & 4357 & 544 & 544 \\ \hline
    \end{tabular}

\end{table}

\textbf{IMU} The accelerometer and gyroscope data are captured by the IMU sensor embedded in the right frame of smart glasses called Aria. The data is originally provided at a resolution of 1000Hz, but considering computational cost, we resample it to 200Hz. \RED{In PAMAP2, IMU data is collected at 100Hz from sensors attached to the subject’s chest. We upsample this data to 200Hz using linear interpolation.}

\textbf{Text} The text data is utilized in the following two formats, Expert Commentary and AAD:
\begin{itemize}
\item Expert Commentary: This includes expert annotations, activity names, and the duration for which the commentary is provided. Since the CLIP model used in our study has a maximum token limit of 77, we truncate sentences exceeding this limit to 77 words.
\item Atomic Action Descriptions (AAD): This follows a format similar to the narrations in the Ego4D dataset. It consists of short descriptions written by non-expert third-party annotators to describe the subject's actions.
\end{itemize}

\textbf{Video} Four GoPro cameras are used to capture third-person view videos from four different angles simultaneously. The original videos are in 4K resolution at 60FPS. We preprocess them to 224x224 resolution at 10FPS. The sliding window is set to 5 seconds.

\textbf{Mocap} The motion capture data, referred to as EgoBodyPose, provides the coordinates of 17 body joints, including the right eye, left shoulder, right wrist, and left ankle, at a frame rate of 10FPS. Missing joint data is filled with zero padding. A sliding window of 5 seconds is applied, and missing intervals within 5 seconds are interpolated using linear interpolation.

\subsection{Evaluation Method} \label{section:tasks}

\begin{figure}[h]
\centering
\includegraphics[scale=0.5]{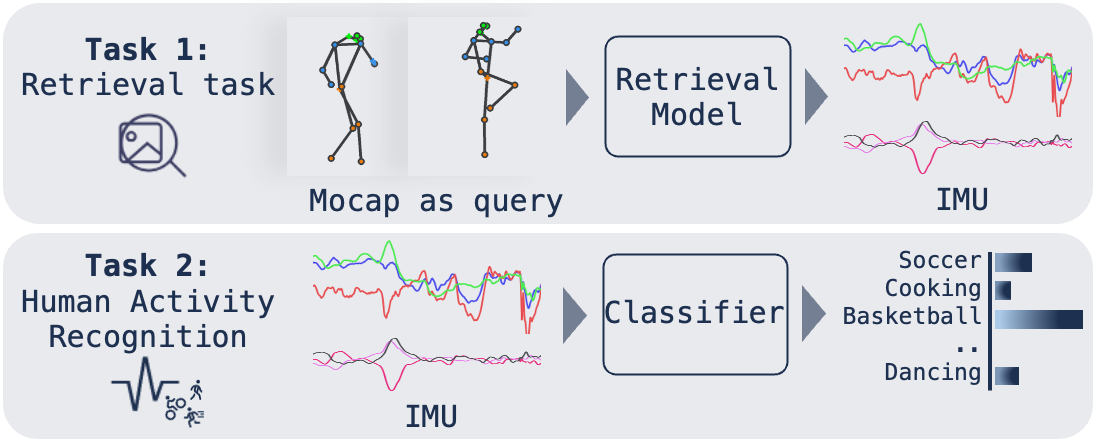}
\caption{Task 1: A retrieval task that measures recall from arbitrary queries to different modalities (top), and Task 2: A human activity recognition task using IMU data (bottom).}
\label{fig:tasks}
\end{figure}

The evaluation framework consists of two primary tasks: Task 1 (Cross-Modal Retrieval) and Task 2 (Human Activity Recognition), as illustrated in Figure \ref{fig:tasks}. Each task is designed to assess different aspects of the model’s performance, ensuring robustness across multiple dimensions of ubiquitous computing applications.

\subsection*{Task1: Cross-Modal Retrieval}
Task 1 evaluates the model's capability to retrieve relevant instances between two arbitrary modalities, emphasizing its effectiveness in cross-modal learning. In this setup, a query-key retrieval paradigm is employed, where a given query from one modality is used to retrieve the most relevant instances from another modality. This is a crucial aspect of ubiquitous computing systems, as they often need to integrate heterogeneous data sources such as IMU, video, motion, and textual descriptions.
The retrieval performance is quantified using Recall@k and Mean Reciprocal Rank (MRR) metrics.
\begin{itemize}
    \item \textbf{Recall@k (R@k)} measures the proportion of relevant items retrieved within the top-k results, which reflects the model's ability to rank correct items higher in the retrieval list.
    \item \textbf{MRR} evaluates the effectiveness of retrieval by considering the ranking of the first relevant item in the search results. It is calculated as the average of the reciprocal ranks across all queries.
\end{itemize}
This study evaluates cross-modal retrieval across four different sensor modalities:
\begin{itemize}
    \item[-] IMU $\leftrightarrow$ Mocap
    \item[-] IMU $\leftrightarrow$ Expert Commentary
    \item[-] IMU $\leftrightarrow$ AAD
    \item[-] IMU $\leftrightarrow$ Video
\end{itemize}
Since retrieval is conducted bidirectionally for each modality pair (e.g., IMU $\rightarrow$ Mocap and Mocap $\rightarrow$ IMU), a total of eight retrieval tasks are evaluated. This setup enables a comprehensive assessment of the model’s ability to integrate and retrieve information across diverse sensor data sources.

\subsection*{Task2: Human Activity Recognition Using IMU Data}
Task 2 focuses on Human Activity Recognition (HAR) based on IMU sensor data. HAR is a crucial area in ubiquitous computing, enabling applications such as health monitoring, smart environments, and activity-based user interfaces. In this task, the model receives a sliding window of raw IMU signals and predicts the corresponding activity labels.
The activity labels are derived through a soft annotation process via text matching, which assigns labels based on semantic similarity between motion patterns and textual descriptions.

\RED{In the Ego-Exo4D dataset, activities are categorized into the following eight labels: "Bike Repair", "Soccer", "Cooking", "Health", "Music", "Rock Climbing", "Basketball", and "Dance". In the PAMAP2 dataset, activities are categorized into the following 18 labels:
"lying", "sitting", "standing", "walking", "running", "cycling", "Nordic walking", "watching TV", "computer work", "car driving", "ascending stairs", "descending stairs", "vacuum cleaning", "ironing", "folding laundry", "house cleaning", "playing soccer", "rope jumping"}
Since classification relies on text-derived labels, we exclude evaluations that use encoders trained solely between Mocap and IMU data. Instead, the focus is placed on evaluating the model’s ability to generalize across different sensor configurations and activity categories.

\subsection*{Discussion and Implications}
The two-task evaluation framework ensures that the proposed model is assessed from both retrieval-based and classification-based perspectives, covering a broad spectrum of ubiquitous computing applications. By integrating cross-modal retrieval and HAR tasks, we demonstrate the model’s ability to handle diverse multimodal data sources, which is essential for building adaptive and scalable ubiquitous systems.

\subsection{Experimental Results} \label{section:result}
We present the results and discussion for Task 1 (retrieval task) and Task 2 (human activity recognition from IMU data). First, we analyze the retrieval task, evaluating how well AURA-MFM aligns different modalities by retrieving the most relevant samples given arbitrary queries. 
We provide results, including retrieval performance for test set queries as well as novel prompts unseen in test set, demonstrating the model’s generalization capabilities.

\begin{table}[h]
  \centering
  \caption{The results of Retrieval tasks}
  \label{tb:retrieval}
  \scalebox{0.85}{
  \begin{tabular}{c|cc|cccc}
    \hline
    \textbf{Modality pairs}&\multicolumn{2}{c|}{\textbf{Models}}& R@1 & R@10 & R@50 & MRR \\
    \hline
    \hline
    \multirow{2}{*}{IMU $\rightarrow$ Mocap} &
        \multirow{2}{*}{AURA-MFM}
        & RNN         & .0250 & .1875 & .5500 & .0849 \\
        & & Transformer & \textbf{.2208} & \textbf{.6666} & \textbf{.8541} & \textbf{.3667} \\
    \hline
        \multirow{2}{*}{Mocap $\rightarrow$ IMU} &
        \multirow{2}{*}{AURA-MFM}
        & RNN         & .0333 & .1875 & .5291 & .0915 \\
        
        & &Transformer & \textbf{.2208} & \textbf{.6541} & \textbf{.8458} & \textbf{.3607} \\
    \hline
    \hline
        \multirow{3}{*}{\scriptsize{IMU} $\rightarrow$ Expert Commentary}
        & \multicolumn{2}{c|}{IMU2CLIP \cite{moon2023imu2clip}} & .0010 & .0149 & .0719 & .0095 \\ \cline{2-3}
        &\multirow{2}{*}{AURA-MFM}
        & RNN         & .0120 & .1030 & .4199 & .0497 \\
        & &Transformer & \textbf{.0199} & \textbf{.1720} & \textbf{.4869} & \textbf{.0710} \\
    \hline
        \multirow{3}{*}{\scriptsize{Expert Commentary} $\rightarrow$ IMU}
        & \multicolumn{2}{c|}{IMU2CLIP \cite{moon2023imu2clip}} & .0020 & .0099 & .0780 & .0093 \\\cline{2-3}
        &\multirow{2}{*}{AURA-MFM}
        & RNN         & .0140 & .1180 & .4629 & .0560 \\
        
        & &Transformer & \textbf{.0309} & \textbf{.1749} & \textbf{.4930} & \textbf{.0819} \\
    \hline
    \hline
        \multirow{3}{*}{IMU $\rightarrow$ AAD}
        & \multicolumn{2}{c|}{IMU2CLIP \cite{moon2023imu2clip}} & .0000 & .0041 & .0402 & .0051 \\ \cline{2-3}
        &\multirow{2}{*}{AURA-MFM}
        & RNN         & \textbf{.0148} & \textbf{.1472} & \textbf{.4679} & \textbf{.0626} \\
        
        & &Transformer & .0106 & .1085 & .4185 & .0510 \\
    \hline
        \multirow{3}{*}{AAD $\rightarrow$ IMU}
        & \multicolumn{2}{c|}{IMU2CLIP \cite{moon2023imu2clip}} & .0008 & .0082 & .0460 & .0063 \\ \cline{2-3}
        &\multirow{2}{*}{AURA-MFM}
        & RNN         & \textbf{.0222} & \textbf{.1562} & \textbf{.4745} & \textbf{.0714} \\
        
        & &Transformer & .0164 & .1233 & .4268 & .0602 \\
    \hline
    \hline
        \multirow{3}{*}{IMU $\rightarrow$ Video}
       & \multicolumn{2}{c|}{IMU2CLIP \cite{moon2023imu2clip}} & .0000 & .0078 & .0453 & .0057 \\  \cline{2-3}
        &\multirow{2}{*}{AURA-MFM}
        & RNN         & \textbf{.0171} & \textbf{.1328} & \textbf{.4367} & \textbf{.0605} \\
        
        & &Transformer & .0101 & .1109 & .4101 & .0505 \\
    \hline
        \multirow{3}{*}{Video $\rightarrow$ IMU}
        & \multicolumn{2}{c|}{IMU2CLIP \cite{moon2023imu2clip}} & .0000 & .0093 & .0390 & .0053 \\\cline{2-3}
        & \multirow{2}{*}{AURA-MFM} 
        & RNN         & \textbf{.0164} & \textbf{.1429} & \textbf{.4734} & \textbf{.0652} \\
        
        & & Transformer & \textbf{.0164} & .1140 & .4109 & .0561 \\
    \hline
    \hline
  \end{tabular}
  }
\end{table}

\subsection*{\B{Result 1: Cross-Modal Retrieval Task}}



Table \ref{tb:retrieval} presents the results of the retrieval task, where different modalities are used as queries to retrieve corresponding data from another modality. The notations shown on the left side of the table, such as IMU $\rightarrow$ Mocap, indicate the recall accuracy when retrieving motion capture data given an IMU instance as input. This retrieval setting evaluates how well the learned multimodal representations align across different sensor modalities. Our proposed AURA-MFM consistently outperforms the baseline model, IMU2CLIP \cite{moon2023imu2clip}, across multiple modalities, including IMU, text, and video.

Among various retrieval scenarios, the most significant improvement is observed in Expert Commentary $\rightarrow$ IMU, where our model achieves an MRR increase of +0.0726. Additionally, retrieval performance metrics improve significantly, with R@1, R@10, and R@50 increasing by +0.0289, +0.1650, and +0.4150, respectively. These results suggest that our approach more effectively captures the underlying relationships between expert commentary and IMU signals compared to previous methods. However, it is important to note that while IMU2CLIP was originally trained on the Ego4D dataset \cite{Ego4D2022CVPR}, our experiments utilize the newly curated Ego-Exo4D dataset \cite{Grauman_2024_CVPR}. Differences in dataset size, sensor modalities, and camera perspectives introduce inherent biases in the test set, which may contribute to performance variations compared to prior benchmarks.
\B{
Specifically, for retrieval tasks involving IMU and text or video modalities, the original paper \cite{moon2023imu2clip} reports Mean Reciprocal Rank (MRR) scores of 0.104 for IMU $\rightarrow$ Text and 0.226 for Video $\rightarrow$ IMU. Additionally, recall metrics such as R@1, R@10, and R@50 were reported as 4.86, 18.75, and 48.26 for IMU $\rightarrow$ Text, and 12.19, 45.31, and 80.00 for Video $\rightarrow$ IMU, respectively. These results are directly cited from their experiments conducted on the Ego4D dataset \cite{Ego4D2022CVPR}.
It is important to note that while these baseline results provide a valuable point of reference, the comparison is not entirely direct due to differences in datasets and experimental setups. Unlike the Ego4D dataset used in the original IMU2CLIP study, our experiments are based on the newly curated Ego-Exo4D dataset \cite{Grauman_2024_CVPR}. Variations in dataset size, sensor modalities, and camera perspectives inevitably introduce some degree of bias, which must be carefully considered when interpreting these findings.
}

Next, we demonstrate that our newly introduced Transformer-based encoder significantly enhances retrieval performance when trained using motion capture and expert commentary data. For the IMU $\rightarrow$ Mocap retrieval setting, our model shows a substantial performance boost over the RNN-based approach ({\sl Stacked 1D-CNN-RNN}), with MRR increasing by +0.2818. Furthermore, recall metrics improve notably, with R@1, R@10, and R@50 increasing by +0.1958, +0.4791, and +0.3041, respectively. Similarly, in the Expert Commentary $\rightarrow$ IMU retrieval setting, our Transformer-based encoder surpasses the RNN-based approach, leading to an MRR improvement of +0.0259 and recall enhancements of +0.0169, +0.0569, and +0.0301 for R@1, R@10, and R@50, respectively. These improvements indicate that the use of Transformer-based architectures enables more effective alignment between IMU and other modalities, particularly for retrieval tasks involving textual descriptions of human activities.

\begin{figure}[h]
\centering
\includegraphics[scale=0.5]{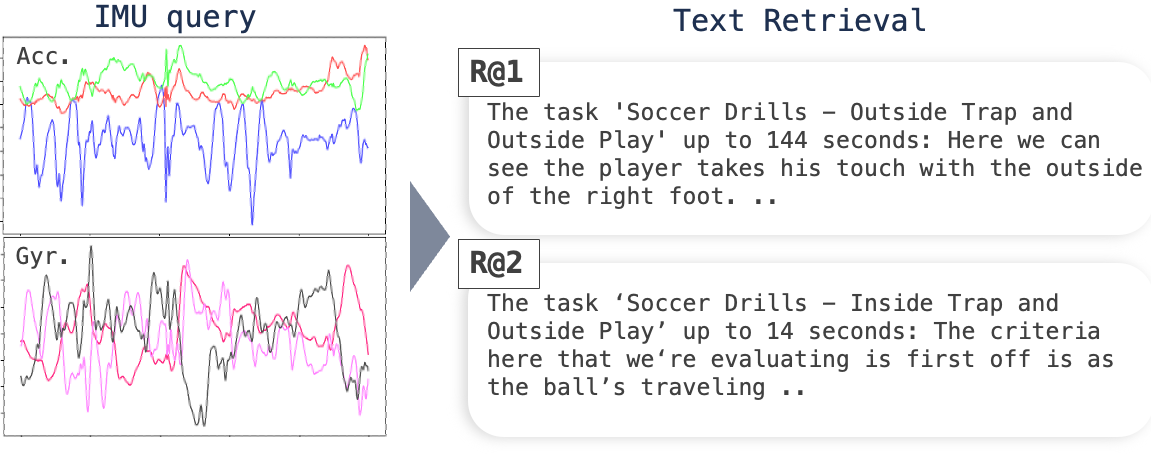}
\caption{\textbf{Retrieval results for IMU$\rightarrow$ Expert Commentary}: \RED{This figure shows the retrieved text (Expert Commentary) from the Ego-Exo4D test set in response to an IMU query. The ground-truth activity label for the IMU query is "Soccer Drills - Outside Trap and Outside Play."} The top-ranked results (R@1 and R@2) accurately capture the passing motion in soccer, demonstrating the effectiveness of our approach in aligning IMU signals with relevant expert commentary.}
\label{fig:retrieval_imu2text}
\end{figure}
\begin{figure}[h]
\centering
\includegraphics[scale=0.55]{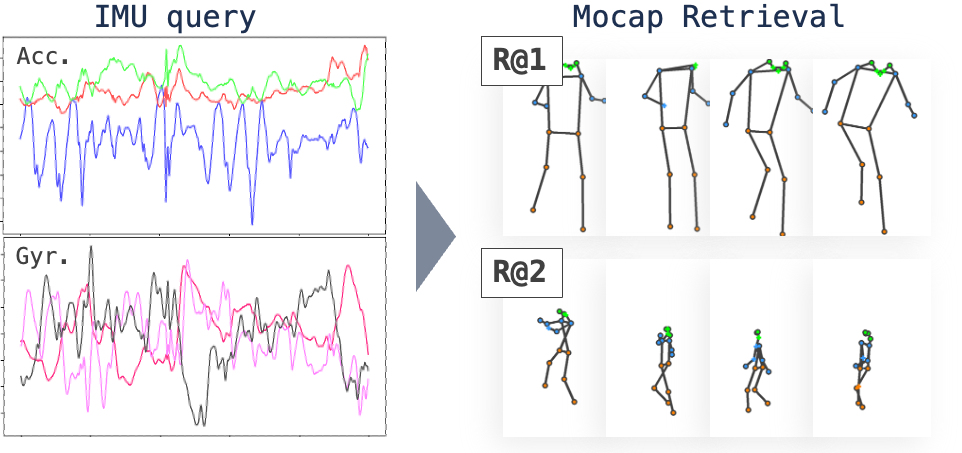}
\caption{\textbf{Retrieval results for IMU $\rightarrow$ Mocap}: The figure shows motion capture sequences retrieved based on IMU signals. \RED{The IMU query is the same as the one shown in Figure 5, and the ground-truth activity label is "Soccer Drills - Outside Trap and Outside Play."} The top-ranked result (R@1) depicts a right-foot trap, while the second-ranked result (R@2) shows a right-foot pass and kick. Both actions align well with the query, demonstrating the model's ability to retrieve precise movement representations. Although visualized from a fixed perspective, the original Mocap data remains three-dimensional, ensuring retrieval accuracy regardless of viewpoint.}
\label{fig:retrieval_imu2mocap}
\end{figure}

Figure \ref{fig:retrieval_imu2text} further illustrates the retrieval results for IMU $\rightarrow$ Expert Commentary. On the left side of the figure, the IMU query consists of accelerometer and gyroscope waveforms corresponding to the activity labeled as "Soccer Drills - Outside Trap and Outside Play." This IMU signal captures the precise motion of a player receiving and passing a soccer ball. The retrieved text descriptions provide additional insight into the quality of the learned multimodal representations. The top-ranked match (R@1) is considered the most relevant, while the second-ranked result (R@2) serves as a closely related alternative. Notably, for text retrieval, both R@1 and R@2 contain descriptions that emphasize a passing motion in soccer, which strongly aligns with the queried IMU sequence. 
\G{In addition, if we look at the details, considering that the original signal consists of two actions, ``the action of receiving a pass in soccer and giving a pass,'' we can see that the latter action of issuing a pass is completed, but the action of receiving a pass is successful. This behavior is difficult to evaluate. In order to say that this action is receiving a pass, I think it is necessary to evaluate this action a little earlier to determine whether or not this action is receiving a pass. Also, if we compare the sentences of R@1 and R@2, there is a difference in whether they are Outside Trap or Inside Trap. The difference between these is whether it is received on the outside of the foot or on the inside of the foot.
Based on this person's skeletal diagram, it seems difficult to determine which of these is the case. In addition, the evaluation was performed a little before this behavior.
Even when I tried this, it seemed difficult to determine whether the ball was received on the outside of the foot or the inside of the foot, and I felt that a different evaluation method was needed for this.}
This qualitative evaluation further supports the effectiveness of our approach in capturing meaningful semantic relationships between IMU data and textual descriptions, reinforcing the broader applicability of AURA-MFM in retrieval-based tasks.

Figure \ref{fig:retrieval_imu2mocap} illustrates the retrieval results for IMU $\rightarrow$ Mocap, where motion capture data is retrieved based on IMU signal input. In the retrieved Mocap sequences, the top-ranked result (R@1) appears to depict a right-foot trap of a soccer ball, while the second-ranked result (R@2) shows a right-foot pass and kick. Since both actions are highly relevant to the query, these retrievals are considered accurate. It is worth noting that although the visualization in the figure presents Mocap data from a fixed camera perspective, the original motion capture data itself is three-dimensional. This means that the evaluation of retrieval performance remains unaffected by the viewing angle used for visualization. The results confirm that our model effectively captures fine-grained movement representations, enabling accurate retrieval based on IMU signals.

\begin{figure}[h]
\centering
\includegraphics[scale=0.36]{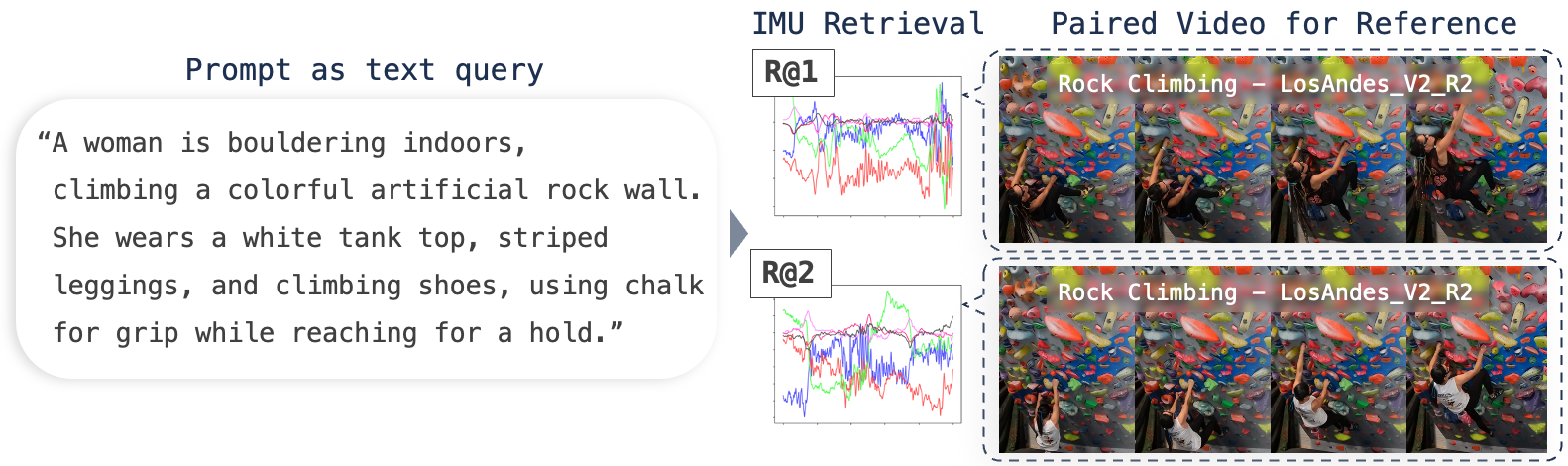}
\caption{\textbf{Retrieval results of IMU from freely created prompts (Prompt $\rightarrow$ IMU)}: \RED{This figure shows the retrieval results for the IMU modality when given a free-form text query that is not included in the test set.
For reference, video frames synchronized with the IMU waveforms are also shown.
The ground-truth activity label for the IMU corresponds to "Rock Climbing - LosAndes V2 R2," demonstrating the model’s ability to generalize IMU waveforms from semantically related textual descriptions, such as "Bouldering," included in the query.}}
\label{fig:retrieval_prompt2imu}
\end{figure}
\begin{figure}[h]
\centering
\includegraphics[scale=0.36]{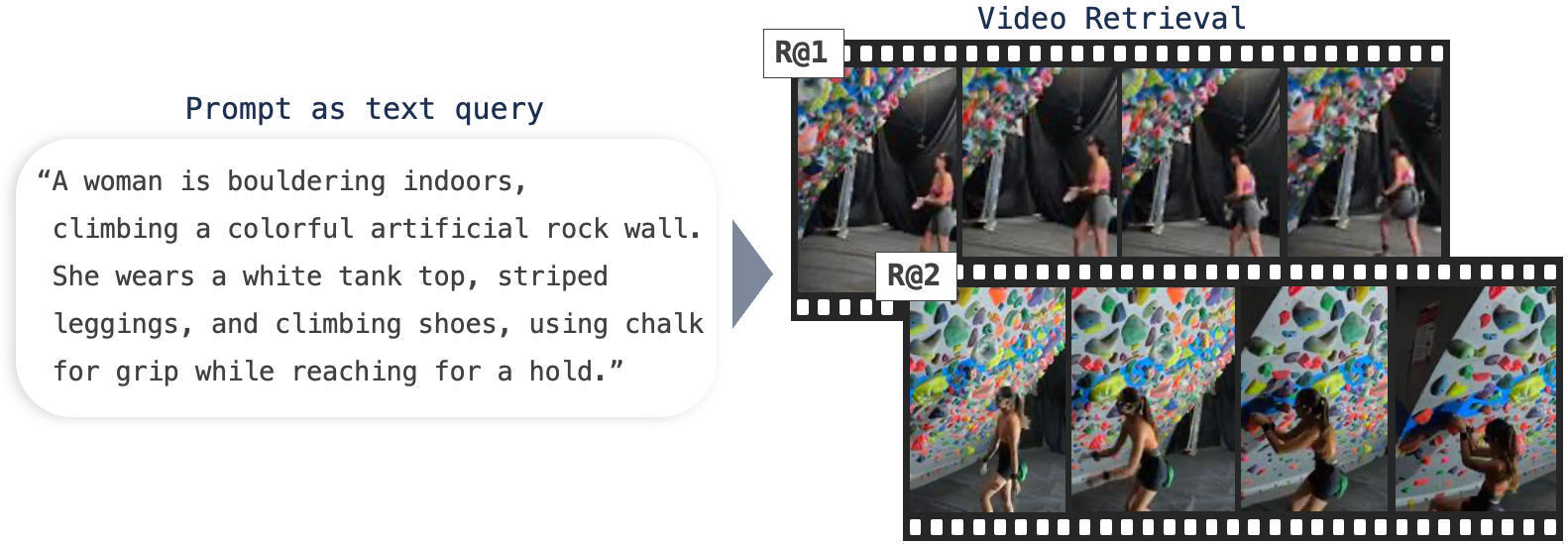}
\caption{\textbf{Retrieval results of Video from freely created prompts (Prompt $\rightarrow$ Video)}: \RED{This figure shows the retrieval results for the video modality when given a free-form text query (identical to Figure 7). The retrieved clips mainly depict scenes related to rock climbing, but they focus on the moments just before the actual climbing begins. As a result, they don't fully align with the present progressive nuance of the prompt phrase 'is bouldering.' While there are some slight differences in clothing, the overall content remains relevant, indicating a certain capability to retrieve related video content even for unknown prompts.}}
\label{fig:retrieval_prompt2video}
\end{figure}

Figure \ref{fig:retrieval_prompt2imu} and Figure \ref{fig:retrieval_prompt2video} show the retrieval results for both IMU and Video modalities in response to a freely composed prompt text\footnote{"A woman is bouldering indoors, climbing a colorful artificial rock wall. She wears a white tank top, striped leggings, and sclimbing shoes, using chalk for grip while reaching for a hold."} that is not present in our test set. Unlike the previous experiments, which evaluated retrieval performance using predefined prompts from the test set, this experiment investigates how well the system can handle an entirely new, unseen prompt.
Although the activity "Bouldering" is dataset-dependent, we have confirmed that this specific description does not appear in the dataset. Regarding the IMU results, while it is inherently challenging to interpret this activity solely from IMU signal waveforms, both R@1 and R@2 are labeled as "Rock Climbing - LosAndes\_V2\_R2". This suggests that the system successfully generalizes from related textual descriptions such as "Bouldering" and "Climbing a rock wall" to IMU waveforms. The figure also includes synchronized video frames alongside the IMU waveforms, providing additional visual context for understanding the retrieved activity.
As for the Video results, both R@1 and R@2 retrieve rock climbing footage. However, the retrieved clips primarily capture moments just before the actual climbing begins, which does not perfectly align with the present progressive nuance of "is bouldering" in the prompt. Additionally, while the woman's attire in the retrieved videos differs slightly from the prompt description, the overall content remains reasonably consistent. Given that the test set does not contain videos that exactly match this prompt, this result demonstrates the system’s capability to retrieve relevant content even for novel, unconstrained inputs.
\begin{table}[t]
  \centering
  \caption{The results of human activity recognition (Ego-Exo4D)}
  \label{tb:har}
  \scalebox{0.8}{
  \begin{tabular}{cccccccc}
    \hline
    \multicolumn{2}{c}{\multirow{2}{*}{\centering \textbf{Models}}} & \multicolumn{2}{c}{\textbf{Zeroshot}} & \multicolumn{2}{c}{\textbf{Transfer learning}} &\multicolumn{2}{c}{\textbf{Fine-tuning}}\\
    \cmidrule(lr){3-4} \cmidrule(lr){5-6}\cmidrule(lr){7-8}
    & & F1 & Acc. & F1 & Acc.& F1 & Acc. \\
    \hline
    \multirow{2}{*}{Random Init.}
        & RNN         & - & - & - & - & .4078 & .6626 \\
        & Transformer & - & - & - & - & .6251 & .7511 \\
    \hline
    \multicolumn{2}{c}{IMU2CLIP \cite{moon2023imu2clip}} & .0747 & .1961 & .4332 & .6315 & .5983 & .7033 \\
    \hline
    \hline
        AURA-MFM
        & RNN         & .5757 & .7057 & .5662 & .7368 & .6228 & .7464 \\
        \scriptsize{(Expert Commentary $\leftrightarrow$ IMU)}
        & Transformer & .6217 & .7081 & .6285 & .7559 & .6491 & .7535 \\
    \hline
        AURA-MFM
        & RNN         & .0805 & .2009 & .3163 & .5263 & .4801 & .6626 \\
        \scriptsize{(AAD $\leftrightarrow$ IMU)}
        & Transformer & \textbf{.6226} & \textbf{.7320} & \textbf{.7217} & \textbf{.8181} & \textbf{.7506} & \textbf{.8277} \\
    \hline
        AURA-MFM
        & RNN         & .1111 & .2057 & .3532 & .4808 & .5219 & .6842 \\
        \scriptsize{(Video $\leftrightarrow$ IMU)}
        & Transformer & .5901 & .7224 & .6966 & .8062 & .7217 & .7990 \\
    \hline
  \end{tabular}
  }
\end{table}
\subsection*{\B{Result 2: Human Activity Recognition Task from IMU Data}}

\begin{table}[t]
  \centering
  \caption{\RED{The results of human activity recognition (PAMAP2)}}
  \label{tb:har_pamap2}
  \scalebox{0.8}{
  \begin{tabular}{cccccccc}
    \hline
    \multicolumn{2}{c}{\multirow{2}{*}{\centering \textbf{Models}}} & \multicolumn{2}{c}{\textbf{Zeroshot}} & \multicolumn{2}{c}{\textbf{Transfer learning}} &\multicolumn{2}{c}{\textbf{Fine-tuning}}\\
    \cmidrule(lr){3-4} \cmidrule(lr){5-6}\cmidrule(lr){7-8}
    & & F1 & Acc. & F1 & Acc.& F1 & Acc. \\
    \hline
    \multirow{2}{*}{Random Init.}
        & RNN         & - & - & - & - & .5908 & .6488 \\
        & Transformer & - & - & - & - & .6548 & .7022 \\
    \hline
    \multicolumn{2}{c}{IMU2CLIP \cite{moon2023imu2clip}} & .0404 & .0404 & \textbf{.6498} & \textbf{.6709} & .6536 & .6893 \\
    \hline
    \hline
        AURA-MFM
        & RNN         &.0151 & .0238 &.5915 & .6084 & .6727& .7077 \\
        \scriptsize{(Expert Commentary $\leftrightarrow$ IMU)}
        & Transformer &.0293 & .0422 &.5070& .5845 & .7222 &  .7389\\
    \hline
        AURA-MFM
        & RNN        &.0293 &.0496 &.3965 &.4926  & .6255 & .6617 \\
        \scriptsize{(AAD $\leftrightarrow$ IMU)}
        & Transformer & \textbf{.0580} & \textbf{.0937} &.3838 &  .4852&\textbf{.7521}  & \textbf{.7738} \\
    \hline
        AURA-MFM
        & RNN         &.0170  & .0294 & .3581 &.4558 & .6590 &.6948 \\
        \scriptsize{(Video $\leftrightarrow$ IMU)}
        & Transformer &.0111 & .0165& .5305 & .5533 &.6983 & .7408 \\
    \hline
  \end{tabular}
 }
\end{table}


Table \ref{tb:har} presents the results of the human activity recognition (HAR) task in Ego-Exo4D dataset, highlighting the performance of different models under three evaluation settings: zero-shot classification, transfer learning, and fine-tuning. The table compares our proposed AURA-MFM framework with the existing IMU2CLIP \cite{moon2023imu2clip} method and baseline randomly initialized models. The evaluation metrics used are F1 score (F1) and accuracy (Acc.), both of which provide insights into the effectiveness of each model in recognizing human activities from IMU data.

In the zero-shot classification setting, our proposed AURA-MFM models significantly outperformed IMU2CLIP. While IMU2CLIP achieved an F1 score of 0.0747 and an accuracy of 0.1961, the best-performing AURA-MFM model, trained with the AAD $\leftrightarrow$ IMU pairing and using a Transformer-based encoder, achieved a much higher F1 score of 0.6226 and accuracy of 0.7320. This substantial improvement suggests that our approach effectively captures meaningful feature representations even in a zero-shot setting, where the model has not been explicitly trained on the target activities. Among different variations of AURA-MFM, the model tuned with Expert Commentary $\leftrightarrow$ IMU also showed strong performance (F1: 0.6217, Acc: 0.7081), indicating the potential benefits of expert-annotated textual descriptions in enhancing zero-shot activity recognition.

Moving to the transfer learning setting, where models are pre-trained on related data before being adapted to the HAR task, we observe further performance gains across all models. IMU2CLIP, which performed poorly in zero-shot classification, saw a notable improvement, achieving an F1 score of 0.4332 and an accuracy of 0.6315. However, AURA-MFM continued to demonstrate superior results. Notably, the Transformer-based AURA-MFM model trained with AAD $\leftrightarrow$ IMU data achieved the highest performance (F1: 0.7217, Acc: 0.8181), showing a significant advantage over both IMU2CLIP and RNN-based variants of AURA-MFM. The RNN-based model, while performing reasonably well in some configurations, did not achieve the same level of consistency, with its best-performing variant (Expert Commentary $\leftrightarrow$ IMU) reaching an F1 score of 0.5662 and accuracy of 0.7368. This result suggests that the Transformer-based encoder is better suited for handling complex multimodal relationships and learning generalizable representations from IMU data.

In the fine-tuning setting, where models are fully trained on labeled IMU data, all methods achieved their highest accuracy. IMU2CLIP reached an F1 score of 0.5983 and accuracy of 0.7033, showing that full supervision helped bridge some performance gaps. However, AURA-MFM again demonstrated superior results, particularly when using a Transformer-based encoder. The best performance was observed in the AURA-MFM model trained with AAD $\leftrightarrow$ IMU (F1: 0.7506, Acc: 0.8277), which significantly outperformed both IMU2CLIP and other AURA-MFM variants. The video-tuned AURA-MFM model also showed strong performance (F1: 0.7217, Acc: 0.7990), confirming the effectiveness of our multimodal learning strategy. The RNN-based models performed comparatively well but did not reach the same accuracy as Transformer-based models, with their best result (Expert Commentary $\leftrightarrow$ IMU) achieving F1: 0.6228 and Acc: 0.7464.

\RED{Table \ref{tb:har_pamap2} presents the results of the human activity recognition (HAR) task in the PAMAP2 dataset. In the zero-shot classification setting, compared to the results in Ego-Exo4D (Table \ref{tb:har}), overall performance was significantly lower. Although the best-performing AURA-MFM model (trained with AAD $\leftrightarrow$ IMU pairing and using a Transformer-based encoder) outperformed the existing IMU2CLIP, its performance remained relatively low. This result is likely caused by domain differences between Ego-Exo4D and PAMAP2. These datasets differ substantially in activities, sampling rates, and sensor placement locations. For example, in terms of activities, the only common activity between PAMAP2 and Ego-Exo4D is "soccer", and PAMAP2 includes a larger number of activities than Ego-Exo4D. Regarding sampling rates, Ego-Exo4D downsamples 1000Hz data to 200Hz, whereas PAMAP2 upsamples 100Hz data to 200Hz through linear interpolation. As for sensor placement, PAMAP2 selected the chest position. While this is closer to the head position used in Ego-Exo4D compared to other placement options like arms or legs, movement differences still appear in the sensor data. In zero-shot classification, which involves no model training, we can see that AURA-MFM faces challenges in generalization capability across datasets with different domains. The results in the transfer learning setting were likely caused by these same factors. However, in the fine-tuning setting, Transformer-based AURA-MFM (AAD $\leftrightarrow$ IMU) achieved F1 scores of .7521 and accuracy of .7738, outperforming IMU2CLIP and the randomly initialized model. Given that Transformer-based AURA-MFM also demonstrated the highest performance in zero-shot classification, this suggests the effectiveness of Transformer-based encoders compared to RNN-based encoders.}

An important observation from these results is the advantage of Transformer-based encoders over RNN-based encoders. While the RNN-based model performed well when trained with Expert Commentary, it struggled in configurations involving AAD and Video data. This suggests that RNNs may be prone to overfitting on smaller datasets, limiting their generalization ability. In contrast, the Transformer-based encoder demonstrated consistently high performance across different configurations. The success of AAD $\leftrightarrow$ IMU tuning in particular can be attributed to the fact that this dataset contained approximately ten times more training data than Expert Commentary, allowing the model to learn more robust and transferable feature representations.


Overall, these results highlight the effectiveness of our proposed AURA-MFM framework in human activity recognition tasks. The combination of Transformer-based encoding, multimodal learning, and text-based classification enables outperforming existing IMU-based methods across all evaluation settings in Ego-Exo4D. Notably, AURA-MFM with fine-tuning in the AAD $\leftrightarrow$ IMU setting achieved the highest performance. These findings suggest that our approach is well-suited for real-world applications where IMU data must be used for accurate activity recognition in both supervised and zero-shot scenarios. \RED{At the same time, while evaluations on downstream tasks using PAMAP2 show that AURA-MFM can surpass existing models under certain conditions, there is still room for improvement in its generalization capability. Since the cross-modal contrastive pretraining uses only the Ego-Exo4D dataset, incorporating multiple datasets from different domains in the pretraining phase could potentially improve generalization performance. However, datasets like Ego-Exo4D—which include synchronized IMU, text, third-person video, and motion capture data—are extremely limited, and this remains one of the current constraints of AURA-MFM.} 
\subsection*{\B{Summary of Results 1 and 2}}
\B{
\textbf{Cross-Modal Retrieval Task}: As summarized in Table \ref{tb:retrieval}, AURA-MFM significantly outperformed IMU2CLIP across all metrics in retrieval tasks. Additionally, when comparing encoder types (RNN vs. Transformer), the Transformer encoder achieved superior performance in four out of the eight retrieval scenarios for R@1, while the RNN encoder excelled in three scenarios, with one resulting in a tie. A similar trend was observed in MRR evaluations. Furthermore, as shown in Figures \ref{fig:retrieval_prompt2imu} and \ref{fig:retrieval_prompt2video}, even for freely constructed prompts not included in the test set, the results reflect the corresponding textual prompts. This indicates the versatility and generalization capability of AURA-MFM.\\
\textbf{Human Activity Recognition Task}: For Human Activity Recognition tasks, AURA-MFM demonstrated consistently superior performance compared to IMU2CLIP across various evaluation settings in Ego-Exo4D. Specifically, in terms of the F1 score, AURA-MFM achieved a remarkable improvement of +55\% under zero-shot settings. In addition, AURA-MFM surpassed IMU2CLIP by +29\% and +16\% under transfer learning and fine-tuning settings, respectively, further highlighting its robust performance across different learning paradigms. \RED{On the other hand, while evaluations on PAMAP2—a dataset with a different domain from Ego-Exo4D—show that AURA-MFM outperforms existing models under certain conditions, its generalization capability still has room for improvement. The highest performance was achieved by AURA-MFM in the AAD $\leftrightarrow$ IMU fine-tuning setting.}
}
\section{Conclusion} \label{section:conclusion}
In this study, we propose AURA-MFM, a novel multimodal foundation model that builds upon the foundation of IMU2CLIP\cite{moon2023imu2clip} by integrating inertial measurement unit (IMU) data with text, third-person perspective videos, and human motion capture data, while also incorporating a Transformer-based IMU encoder.
While prior research has predominantly focused on first-person perspective videos, our work underscores the critical importance of incorporating third-person viewpoints and motion capture data to achieve a more holistic, comprehensive, and detailed analysis of full-body human movements. This is crucial because first-person perspectives alone are inherently limited in their ability to fully capture the nuances of complex human actions. Furthermore, we employed a Transformer-based IMU encoder, which has gained significant attention in recent years, and empirically confirmed its effectiveness compared to a traditional RNN-based IMU encoder to a certain degree in our experimental settings. Our model was evaluated through retrieval and activity recognition tasks, and the results demonstrate that our proposed method surpasses the performance of the prior method. Specifically, in zero-shot classification for activity recognition tasks, all variants of our proposed AURA-MFM consistently and significantly outperformed the prior model in Ego-Exo4D.
Notably, our best-performing model achieved an F1-score of 0.6226 and an accuracy of 0.7320, compared to the prior model's F1-score of 0.0747 and accuracy of 0.1961, showcasing a substantial improvement.

\B{The further avenue of this paper consists of the following three. The first one is to enlarge the downstream tasks. We showed two downstream tasks of this multi-modal foundation model: the retrieval and the activity recognition tasks. The variety of ubuquitous tasks such as activity localization tasks for IMU / WIFI signals, smart house task are among this category of downstream tasks. 
The second one would be related to the characteristics of this multimodal foundation model. As is mentioned in the multimodal dataset in the past in the subsection of Ego-Exo4D dataset, they typically do not contain the text modality. This text modality may be very important for some category of tasks.
For example, the medical tasks such as the prediction of Alzheimer disease and stress recognition tasks have text data as well.  
However, these tasks are often built on the dataset with single modality of signals despite that they have text data such as electronic form of assessment of medical doctor. The text data is often easily acquired in the progress of Alzheimer disease or other disease. Similarly, the image data is often acquired as well.
The multimodal foundation model can enlarge this possibility to handle the signal data with other modalities of data: image and text data.} 

\B{The third avenue is related to the differnt training method.
We believe that approaches for optimizing integration methods between each modality and understanding complex movements are also possible. The approach in this paper limited itself with training to one-on-one cross-modal learning. A method that aims to build a multimodal model that integrates four or more modalities and can train them “simultaneously” is a training method that more effectively utilizes the interaction between modalities. In this case, it can be expected that it will work effectively for multimodal data sets where there is strong interaction between modalities.}

\bibliographystyle{plain}
\bibliography{bibtex}

\end{document}